\documentclass[10pt,journal,compsoc]{IEEEtran}
\usepackage{graphicx}
\usepackage{color}
\usepackage{amsmath,url}
\usepackage{comment}
\usepackage{placeins}

\usepackage[utf8]{inputenc}
\usepackage{graphicx} 
\usepackage{booktabs}
\usepackage{array}
\usepackage{tabularx}
\usepackage{multirow}
\usepackage{microtype}
\usepackage{amssymb}
\usepackage[table,xcdraw]{xcolor}

\ifCLASSOPTIONcompsoc
  \usepackage[nocompress]{cite}
\else
  \usepackage{cite}
\fi
\ifCLASSINFOpdf
\else
\fi

\usepackage{booktabs}

\title{Predicting emotion from music videos: exploring the relative contribution of visual and auditory information to affective responses}
\author{Phoebe~Chua,     
        Dimos~Makris,~
        Dorien~Herremans,~\IEEEmembership{Senior Member,~IEEE,}
        Gemma~Roig,~\IEEEmembership{Member,~IEEE}
        and~Kat~Agres,~\IEEEmembership{Member,~IEEE}
\IEEEcompsocitemizethanks{
\IEEEcompsocthanksitem P. Chua is with the National University of Singapore, Singapore.\protect\\ Email: \protect\url{pchua@nus.edu.sg}
\IEEEcompsocthanksitem K. Agres is with the Yong Siew Toh Conservatory of Music, National University of Singapore, Singapore, 117376.\protect\\
E-mail: \protect\url{katagres@nus.edu.sg} 
\IEEEcompsocthanksitem D. Herremans and D. Makris are with Singapore University of Technology and Design, Singapore.
\IEEEcompsocthanksitem G. Roig is with the Goethe University Frankfurt, Germany.}

\thanks{Manuscript received 000; revised 000.}}

\usepackage{booktabs, multirow}
\begin{document}
\IEEEtitleabstractindextext{%
\begin{abstract}

Although media content is increasingly produced, distributed, and consumed in multiple combinations of modalities, how individual modalities contribute to the perceived emotion of a media item remains poorly understood. In this paper we present MusicVideos (MuVi), a novel dataset for affective multimedia content analysis to study how the auditory and visual modalities contribute to the perceived emotion of media. The data were collected by presenting music videos to participants in three conditions: music, visual, and audiovisual. Participants annotated the music videos for valence and arousal over time, as well as the overall emotion conveyed. We present detailed descriptive statistics for key measures in the dataset and the results of feature importance analyses for each condition. Finally, we propose a novel transfer learning architecture to train Predictive models Augmented with Isolated modality Ratings (PAIR) and demonstrate the potential of isolated modality ratings for enhancing multimodal emotion recognition. Our results suggest that perceptions of arousal are influenced primarily by auditory information, while perceptions of valence are more subjective and can be influenced by both visual and auditory information. The dataset is made publicly available.

\end{abstract}

}

\maketitle

\IEEEdisplaynontitleabstractindextext

%
\IEEEpeerreviewmaketitle

\section{Introduction}\label{sec:introduction}
\IEEEPARstart{H}{ow} do we perceive, integrate, and interpret affective information that is conveyed through multiple sensory modalities simultaneously, and how can we model and predict the effects of visual and auditory information on perceived emotion states? The strong connection between music and emotions~\cite{zentner2008emotions} as well as video and emotions~\cite{baveye2013large} is well-studied and intuitive -- indeed, research has found that many people cite emotion regulation as the primary reason they listen to music~\cite{lonsdale2011we}. As multimedia content becomes easier to create and access, music can increasingly be used as a tool to shape the emotional narratives in video~\cite{cohen2010music}; likewise, visual displays can shape our perception of a melody's affective content~\cite{boltz2009audiovisual}. The development of affective multimedia content analysis methods to predict and understand the influence of different media modalities on emotion is hence a key step in the development of affect-aware technologies and has utility in a wide range of applications including content delivery, indexing and search~\cite{lamere2008social}.

The defining feature of multimedia content is its use of multiple modalities to convey information: for example, we might easily imagine a video that plays text (captions), auditory (voices and music) and visual information simultaneously. Publishers have begun to exploit the flexibility of multimedia content by releasing their work on multiple platforms, each tailored to a different modality. For example, contemporary podcasts are frequently recorded with video then released on both audio and video streaming platforms. In turn, users can choose to consume multimedia content in various modalities or combinations of modalities based on their needs and preferences~\cite{chung2015college}. Of the various forms of multimedia content available, music-related content is consumed in an especially wide range of modalities. It can be accessed through audio-only platforms or in audio-visual form (music videos), and is frequently used as background music for videos watched on mute, just to name a few scenarios. Although music is typically understood as an ``art of sound''~\cite{davies1994musical}, video and audio music streaming currently account for similar proportions of the total music streaming volume, with the International Federation of the Phonographic Industry (IFPI) reporting in their latest Global Music Report~\cite{ifpi_2020} that video streaming made up 47\% of music streaming consumption globally in 2019.

While a great deal of work has been dedicated to music and video emotion recognition, to our knowledge, few studies have acknowledged that multimedia content can be flexibly consumed in different combinations of modalities, which may have differing effects on the emotions of the user. In this paper we present \textbf{MuVi}, a novel dataset for affective multimedia content analysis which contains music videos annotated with both dimensional and discrete measures of emotion in three conditions: audiovisual (original music videos), music (music-only) and visual (muted music videos). Each annotation is accompanied by annotator metadata relevant to the perception of emotion in music including gender, song familiarity and musical training. We provide a comprehensive descriptive analysis of the dataset's main features and examine whether modality and demographic factors influence the perceived emotion of stimuli. Then, we investigate the relative contribution of audio and visual modalities to prediction of arousal and valence in audiovisual stimuli, and discuss the most important auditory and visual features. Finally, we propose a novel model architecture that leverages knowledge transfer from isolated modality ratings for multimodal emotion recognition - a Predictive model Augmented with Isolated modality Ratings (PAIR).

The paper is organized as follows: Motivation for this study and related work on emotion recognition systems for media content are covered in Section~\ref{sec2}. Data collection procedures and key features of the MuVi dataset, including dataset analysis and visualization, are covered in Sections~\ref{sec3} and ~\ref{sec4}. In Section ~\ref{sec5}, we present PAIR, our novel architecture for multimodal emotion recognition along with implementation details. Sections~\ref{sec6} and \ref{sec7} detail the experimental setup and evaluation of our approach. Finally, in Section~\ref{sec8}, we conclude and discuss some avenues for future work.

\section{Background}\label{sec2}

\subsection{Models of emotion}
Emotion is generally defined as a collection of psychological states that include subjective experience, expressive behavior and peripheral physiological responses~\cite{gross2011emotion}. 
While there are multiple ways to understand and measure emotions, the two most frequently used in affective computing research are the dimensional~\cite{russell1980circumplex} and categorical~\cite{ekman1992argument} approaches. The \textit{dimensional} approach represents emotions as points in a continuous space defined by a set of affective dimensions. Intuitively, it implies that emotion can be understood as a linear combination of the dimensions. A commonly employed dimensional model for Music Emotion Recognition (MER) is the circumplex model proposed by Russell~\cite{russell1980circumplex}, which consists of two dimensions: arousal (indicating emotional intensity) and valence (indicating pleasant versus unpleasant emotions). This valence-arousal (VA) model is easily applied across different domains and can be used to collect both static and dynamic emotion annotations - the difference being that static annotations generally reflect the overall emotion or mood conveyed by a media item, while dynamic annotations reflect the emotions conveyed by a media item as it unfolds over time. However, the VA model can lack the coherence of categorical approaches that define emotions with semantic terms~\cite{cowen2017self}. 

\textit{Categorical} approaches represent emotions with a set of discrete labels such as ``joy'' or ``fear'', based on the underlying hypothesis that there are a finite number of distinct basic emotions~\cite{ekman1992argument}. Basic emotion theories typically hypothesize that there are between 6 to 15 emotion categories, while recent work suggests that there are up to 27 distinct varieties of general emotional experiences~\cite{cowen2017self}. Zentner et al.~\cite{zentner2008emotions} proposed the Geneva Emotional Music Scale (GEMS) as a domain-specific model of emotion to more accurately capture the range of aesthetic emotions that can be induced by music. Compared to mainstream emotion models, most of the emotions in GEMS are positive. The model also contains categories that are not always included in traditional approaches to the classification of emotions, such as ``transcendence'' and ``nostalgia,'' which help to better capture the range of emotion states induced by music. The full GEMS-45 scale contains 45 terms, but shorter variants of the scale (GEMS-25 and GEMS-9) were later defined through factor analysis. Categorical approaches may be fairly rich, due to the abstract and complex concepts expressed by natural language; they may also be more intuitive, especially when we want to describe the overall emotions conveyed by a song or video. However, they often do not lend themselves readily to dynamic emotion annotations. 

To integrate dimensional and categorical approaches to understanding and measuring emotion, some have suggested that discrete emotions represent positions in an affective space defined by dimensions such as arousal and valence~\cite{russell2003core, paltoglou2012seeing}. Empirical tests of this relationship with emotionally evocative videos~\cite{cowen2017self} have found that affective dimension judgements are able to explain the variance in categorical judgements and vice versa, and that categories seem to have more semantic value than affective dimensions. However, to our knowledge, similar studies to investigate the relative contributions of music and video content on emotion induction have not been conducted.

\subsection{Perception of multimedia content}
An understanding of how we weigh and integrate affective information from different sensory modalities can inform the design of computational models for multimedia content analysis. At the same time, computational models can provide hints as to how humans perceive affective multimedia content~\cite{kim2010music}. With regards to music multimedia content in particular, we are interested in the relative influence of audio and visual input on the perceived emotion of the content. Empirical studies on the influence of visual information on the evaluation music performance have consistently replicated significant effects~\cite{platz2012eye,tsay2013sight}, suggesting that visual information influences the way we perceive sound and music.

More broadly, studies have examined the ways in which music influences the perception of visual scenes, as well as the reverse relationship – the ways in which visual information influences the cognitive processing of music. Boltz~\cite{boltz2001musical} proposed that music influences visual perception by providing an interpretative framework for story comprehension. For example, mood-congruent music could be utilized by artistic directors to heighten emotional impact or clarify the meaning of ambiguous scenes in a visual story, while mood-incongruent music could be used to convey subtler meanings such as irony~\cite{boltz2009audiovisual}. Although comparatively little work has been done to investigate the way visual information influences the interpretation of music content, empirical findings suggest that music videos help maintain listener's interest when songs are relatively ambiguous~\cite{goldberg1993music}. Both perspectives hint at how the construction of a meaningful narrative is an important component of the way we consume media. Hence, one way to think about multimedia might be as a form of expression that provides multiple information channels through which artists can convey meaning to their audience, and conversely by which audiences can make inferences about the artists' intent. Indeed, Sun and Lull~\cite{sun1986adolescent} argue that one of the main appeals of music videos is that the visual scenes enrich a song's meaning and underlying message.

\sloppy

\begin{table*}[ht] \footnotesize
\centering
\begin{tabular}{lcccccccc} 
\toprule
\multirow{2}{*}{Name} & \multicolumn{3}{c}{Domain} & \multicolumn{2}{c}{Emotion measurement} & \multirow{2}{*}{\# items} & \multirow{2}{*}{\begin{tabular}[c]{@{}c@{}}\# annotations \\per item\end{tabular}} & \multirow{2}{*}{Features} \\ 
\cmidrule{2-6}
 & Music & Visual & AV & Static & Dynamic &  &  &  \\ 
\toprule
AMG1608~\cite{chen2015amg1608} & \checkmark &  &  & \begin{tabular}[c]{@{}c@{}}Valence, \\Arousal\end{tabular} &  & 1,608 & 15-32 & \begin{tabular}[c]{@{}c@{}}Data-based (MFCC, tonal, \\spectral, temporal)\end{tabular} \\ 
\cmidrule(r){1-9}
DEAM~\cite{aljanaki2017developing} & \checkmark &  &  &  & \begin{tabular}[c]{@{}c@{}}Valence, \\Arousal\end{tabular} & 1,802 & 5-10 & \begin{tabular}[c]{@{}c@{}}Annotator metadata (mood, \\familiarity, liking, personality \\etc.), source music data\end{tabular} \\ 
\cmidrule(lr){1-9}
Emotify~\cite{aljanaki2016studying} & \checkmark &  &  & GEMS &  & 400 & 16-48 & \begin{tabular}[c]{@{}c@{}}Annotator mood, metadata \\(age, gender, mother \\tongue), liking\end{tabular} \\ 
\cmidrule(lr){1-9}
Moodswings~\cite{schmidt2011modeling} & \checkmark &  &  &  & \begin{tabular}[c]{@{}c@{}}Valence, \\Arousal\end{tabular} & 240 & 7-23 & \begin{tabular}[c]{@{}c@{}}Data-based (MFCC, spectral), \\Echo Nest Timbre features\end{tabular} \\ 
\cmidrule(lr){1-9}
FilmStim~\cite{schaefer2010assessing} &  & \checkmark &  & \begin{tabular}[c]{@{}c@{}}Arousal, \\etc.\end{tabular} &  & 70 & 44-56 & Source video data \\ 
\cmidrule(lr){1-9}
LIRIS-ACCEDE~\cite{baveye2015deep} &  & \checkmark &  & \begin{tabular}[c]{@{}c@{}}Valence, \\Arousal\end{tabular} & \begin{tabular}[c]{@{}c@{}}Valence, \\Arousal\end{tabular} & 9,800 & \begin{tabular}[c]{@{}c@{}}NA, (rankings \\provided)\end{tabular} & Source video data \\ 
\cmidrule(r){1-9}
DEAP~\cite{koelstra2011deap} &  &  & \checkmark & \begin{tabular}[c]{@{}c@{}}Valence, \\Arousal, \\Dominance\end{tabular} &  & 120 & 14-48 & \begin{tabular}[c]{@{}c@{}}EEG data, frontal face video \\of annotators, links to MVs\end{tabular} \\ 
\cmidrule(r){1-9}
RAVDESS~\cite{livingstone2018ryerson} & \checkmark & \checkmark & \checkmark & \begin{tabular}[c]{@{}c@{}}Basic emotions, \\intensity\end{tabular} &  & 2,452 & 10 & Source recordings \\ 
\cmidrule(lr){1-9}
MuVi (the present dataset) & \checkmark & \checkmark & \checkmark & GEMS & \begin{tabular}[c]{@{}c@{}}Valence, \\Arousal\end{tabular} & 81 & 5-9 & \begin{tabular}[c]{@{}c@{}}Data-based (audio, visual), \\annotator metadata (gender, \\familiarity, musical training etc.)\end{tabular} \\
\bottomrule
\end{tabular}
\vspace{0.1cm}
\caption{Comparison of existing datasets for emotion recognition based on music, visual and audiovisual stimuli.}
\label{t:datasets}
\end{table*}

\subsection{Datasets for affective multimedia content analysis}
The development of computational models for emotion recognition and understanding has motivated the collection of novel datasets, usually based on audio, visual, or audiovisual stimuli that are accompanied by (dynamic) time-series annotations of emotion or (static) emotion labels.  While there are many dimensions along which to label affective computing datasets, here we focus on four key factors: emotion measurement, temporal nature, domain and features. 

\textit{Emotion measurement} refers to the categorical and dimensional approaches described earlier, while \textit{temporal nature} refers to whether the modeling target is a single, static label capturing the overall emotion of the content, or a vector that captures the dynamically changing emotion as it unfolds over time. Across datasets, emotion is most frequently measured using continuous valence and arousal scales, though some datasets include additional dimensions such as dominance \cite{koelstra2011deap} or use categorical measures \cite{aljanaki2016studying}. 

\textit{Domain} refers to the modality in which stimuli are annotated; here, we focus on music, visual and audiovisual stimuli. Datasets using music-based stimuli have largely focused on music consumed in an audio-only format~\cite{aljanaki2017developing},~\cite{yang2012machine},~\cite{yang2018review}. One notable exception is the Database for Emotion Analysis using Physiological Signals (DEAP) which used music videos as the underlying stimuli~\cite{koelstra2011deap}, although the focus of the dataset is predicting emotion labels from a combination of data-based features extracted from the underlying stimuli and physiological data. 
Datasets based on multimodal stimuli include those based on film repositories~\cite{baveye2017affective}~\cite{baveye2015deep}, videos collected ``in the wild''~\cite{ringeval2013introducing},~\cite{ong2019modeling} and those provided by the Audiovisual Emotion Challenge (AVEC). We draw particular attention to the Ryerson Audio-Visual Database of Speech and Song (RAVDESS) \cite{livingstone2018ryerson}, which consists of spoken and song utterances vocalized by professional actors to express emotions based on Ekman’s theory of basic emotions \cite{ekman1992argument}. In contrast to the other datasets, the stimuli in RAVDESS were presented and annotated in three conditions: audio-only, video-only and audiovisual.

\textit{Features} that can be used for emotion recognition are also a key component of each dataset. Reviews have summarized extracted features relevant to affect detection in the audio modality such as intensity (loudness, energy), timbre (MFCC) and rhythm (tempo, regularity) features~\cite{yang2018review} and video modality such as colour, lighting key, motion intensity and shot length~\cite{poria2017review},~\cite{wang2015video}. Features that can capture complex latent dimensions in the data, such as the audio embeddings generated by the VGGish model \cite{gemmeke2017audio} \cite{thao2021attendaffectnet}\cite{attendaffect2021}, are also becoming increasingly popular. Features may be provided in the dataset or extracted from the source data if it is available. Apart from data-based features, however, the subjective nature of emotion perception suggests that it may be useful  to build personalized models that account for characteristics of the user in addition to only the characteristics of the media content being consumed~\cite{yang2008regression}~\cite{wang2012personalized}. Several of the datasets in Table~\ref{t:datasets} provide demographic and contextual information such as the annotator's age, gender, familiarity with the media and current mood. 

A comprehensive review of databases that utilize only one modality or other combinations of modalities is out of the scope of this paper, but we refer interested readers to recent survey studies (e.g., \cite{poria2017review} for multimodal emotion recognition, \cite{yang2012machine} for music and \cite{baveye2015liris} for audiovisual). Table 1 instead highlights several of the prominent datasets relevant to music and audiovisual emotion recognition in order to show that affective computing datasets are typically focused on a single domain, restricting a researchers’ ability to study the relative contribution of individual modalities to multimodal emotion recognition. 

The similarities in the literature on music and video emotion recognition present a clear opportunity to bring together these two lines of research and better understand how individuals respond emotionally to media, and specifically music, presented in multimodal formats. To our knowledge, the dataset collected in this research, the MuVi dataset, is the first published dataset that contains multimodal stimuli annotated with both static and time-series measures of emotion in individual, isolated modalities (music, visual) as well as the original multimodal format (audiovisual). In other words, we have three sets of stimuli: the original music videos (audiovisual modality), the music clips (music modality) and muted video clips (visual modality). The presence of both static and  dynamic annotations for each stimuli provides more multifaceted information about its affective content and enables us to study how discrete emotion labels might map onto continuous measures of valence and arousal~\cite{paltoglou2012seeing}. Finally, we provide the anonymized profile and demographic information of annotators. 

MuVi has several advantages over RAVDESS, the only other dataset we are aware of that contains isolated modality ratings. Firstly, while RAVDESS uses short utterances lasting several seconds as stimuli, MuVi uses longer 60s excerpts from music videos. As such, while the stimuli in RAVDESS are only rated with static emotion labels, the stimuli in MuVi are accompanied by both static overall emotion labels and dynamic time-series annotations. Additionally, while a media item is only shown once per participant in MuVi (ignoring modality), it is unclear whether a participant would be exposed to the same utterance in different modalities in RAVDESS - a concern since repeated presentation might influence the perceived emotion of an utterance \cite{livingstone2018ryerson}.

\section{Dataset collection}\label{sec3}
In this section, we describe the experimental protocol for the collection of MuVi dataset. 

\subsection{Stimuli selection}
The MuVi dataset consists of music videos (MVs) annotated in three modalities: music-only, visual (i.e. muted video only), and audiovisual. We selected a corpus of 81 music videos from the LAKH MIDI Dataset~\cite{raffel2016learning}, a collection of MIDI files matched to entries in the Million Song Dataset~\cite{bertin2011million}. Entries in the Million Song Dataset contain additional information about each music video including audio features and metadata, making it a valuable resource for users of MuVi who may be interested in obtaining additional features. As there were some MVs in which the song was preceded by a silent video intro, we manually scanned each MV to determine the timestamp at which music began playing, then obtained excerpts by taking the first minute of each MV beginning from that timestamp.

\subsection{Procedure}
We recruited 48 participants (31 males) aged between 19 and 35 (mean = 23.35y, std = 3.28y). All participants provided informed consent according to the procedures of the SUTD Institutional Review Board. The study was approved by the Institutional Review Board under SUTD-IRB 20-315. Prior to the experiment, participants filled out a survey that included demographic questions and music listening preferences. Once in the lab, participants were informed of the listening experimental protocol and the meaning of the arousal-valence scale used for annotations. For each quadrant of the arousal-valence space, participants could listen to a short (14s-17s) music clip that was typically rated as belonging to that quadrant. An experimenter was present to answer any questions. 

Each participant was then given two practice trials to familiarize themselves with the annotation interface before beginning the actual task. During the session, each participant annotated between 30 to 36 media items (not inclusive of the practice trials) in a randomly chosen media modality, and to ensure the reliability of ratings, the same media item would not be shown to a participant twice even if in a different modality. We compensated participants with a \$10sgd voucher at the end of the session. The final dataset contains a total of 1,494 annotations (music = 350, visual = 349, audiovisual = 342). Each one-minute excerpt received between five to nine annotations in each modality, with the median number of annotations for each media item being six.

\subsubsection{Annotation process}\label{sec3.3}
\begin{figure}[ht]
\centering
    \includegraphics[scale=0.8]{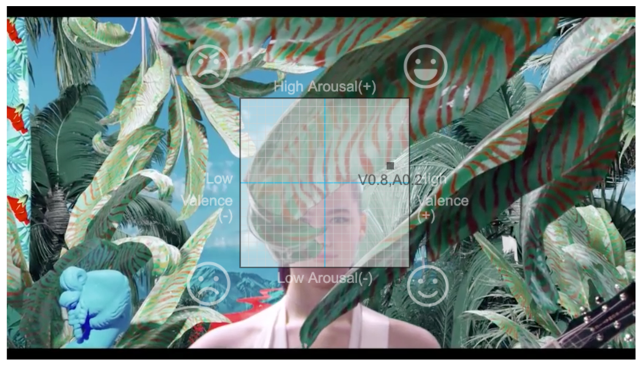}
    \caption{Dynamic annotation interface (overlayed mode). Participants moved their cursor, indicated by the black square, to match the arousal and valence of media items during playback. The interface also provided information about their current coordinates in the 2D space. }
\label{fig:annotation_interface}
\end{figure}

For each media item, participants first completed the dynamic (continuous) annotation task. To date, continuous annotations of arousal and valence for video stimuli have typically been collected using a slider placed outside the video frame~\cite{ong2019modeling, ringeval2013introducing}. However, this design might result in participants having to repeatedly shift their attention between the stimuli and their current position on the slider. Superimposing the arousal-valence scale over the video stimuli might mitigate this issue. To investigate which design participants would prefer, we tested two versions of the dynamic annotation interface in the video-only and music video conditions: with the 2D valence-arousal axes overlaid over the video (Figure~\ref{fig:annotation_interface}) or located beside the video. Each version was randomly selected with $50\%$ probability. Participants were asked to indicate which annotation interface they preferred at the end of the experiment.

For the dynamic annotation task, participants were asked to rate the emotion they thought the media item was trying to convey (rather than how they felt as they watched or listened to the media item) by moving their mouse over a 2D graph of the arousal-valence space as the item played. Figure~\ref{fig:annotation_interface} shows an example of our annotation interface, which displays the axes labels and an emoticon for each quadrant of the arousal-valence space to help participants remember where different perceived emotions lie on the axes.

To collect the dynamic annotations, we used a script that sampled a participant's cursor location every 0.5 seconds. However, the actual sampling frequency slightly varied depending on the lab computer's browser, Internet connection speed and CPU load. For consistency, we resampled the annotations to maintain a sampling interval of as close to 2Hz as possible; the published annotations have a mean sampling interval of 2Hz and a standard deviation of 0.0177s.

\begin{table}[ht!]\small
\begin{tabular}{ lll } 
 \toprule
 \textbf{Superfactor} & \textbf{Category} & \textbf{Emotion labels} \\ 
 \midrule
 Sublimity & Wonder & \begin{tabular}[c]{@{}l@{}}Moved, Allured, \\ Filled with wonder \end{tabular} \\ 
 & Transcendence & \begin{tabular}[c]{@{}l@{}} Fascinated, Overwhelmed, \\ Feeling of transcendence \end{tabular}\\ 
 & Peacefulness & Serene, Calm, Soothed \\ 
 & Tenderness & \begin{tabular}[c]{@{}l@{}}Tender, Affectionate, \\ Mellow \end{tabular}\\ 
 & Nostalgia & \begin{tabular}[c]{@{}l@{}}Nostalgic, Sentimental, \\ Dreamy \end{tabular}\\ 
 \midrule
 Vitality & Power & \begin{tabular}[c]{@{}l@{}}Strong, Energetic, \\ Triumphant \end{tabular}\\
 & Joyful Activation & Animated, Bouncy, Joyful \\
 \midrule
 Unease & Sadness & Sad, Tearful, Blue \\
 & Tension & Tense, Agitated, Nervous \\
 \bottomrule
\end{tabular}
\vspace{0.1cm}
\caption{GEMS-28 emotion terms, together with their categories and superfactors, as per \cite{zentner2008emotions}}
\label{t:discrete_emotions}
\end{table}

Once the media item finished playing, participants indicated whether or not they had watched or listened to the media item before and completed the static (discrete) annotation task. The static annotation task consisted of selecting the terms that they felt described the media item's overall emotion. The terms were taken from GEMS-28 \cite{lykartsis2013emotionality}, an extended version of the GEMS-25 scale, resulting in a total of 27 possible emotion labels. As shown in Table~\ref{t:discrete_emotions}, the labels can be grouped into nine different categories which in turn can be condensed into three ``superfactors''. Both these categories and superfactors were also displayed to participants. Participants had to select at least one label per media item with no upper limit on the number of labels they could choose; the median number of labels chosen was four.

\subsection{Feature extraction}
For each media item in our dataset, we extracted a set of audio and video features. To match the rate at which dynamic arousal-valence annotations were collected, time-varying acoustic features were extracted from the underlying music videos in non-overlapping 500ms windows. 

Audio feature extraction was performed with openSMILE~\cite{eyben2010opensmile}, a popular open-source library for audio feature extraction. Specifically, we used the ``emobase'' configuration file to extract a set of 988 low-level descriptors (LLDs) including MFCC, pitch, spectral, zero-crossing rate, loudness and intensity statistics, many of which have been shown to be effective for identifying emotion in music~\cite{zhang2017feature, cheuk2020regression, thao2021attendaffectnet, attendaffect2021}. Many other configurations are available in openSMILE but we provide the ``emobase'' set of acoustic features since it is well-documented and was designed for emotion recognition applications~\cite{eyben2013recent}.

Six types of visual features are provided: color, lighting key, facial expressions, scenes, objects, and actions. With the exception of action features, visual features were extracted from still frames sampled from the underlying music videos at 500ms intervals. 
Features related to color~\cite{rasheed2002movie} and lighting~\cite{rasheed2005use} have been used in a number of video emotion recognition studies, based on the insight that these visual elements are often manipulated by filmmakers to convey a chosen mood or induce emotional reactions. These insights are supported by experiments showing that the hue and saturation of a color has significant influence on induced arousal and valence~\cite{wilms2018color}. Similarly, there is evidence that the facial expressions of individuals in a scene influence the emotion perceived by a viewer~\cite{li2016happiness}. We use \textbf{OpenCV} \footnote{\url{https://opencv.org}} to extract hue and saturation histograms as well as ``lighting key'' information. Facial expressions were extracted using a pre-trained VGG19~\cite{simonyan2014very} model. 

Recent techniques have begun to recognize the importance of contextual information such as scenes, objects and actions for image and video emotion recognition \cite{chen2016emotion} \cite{lee2019context}. Intuitively, certain contexts may tend to be associated with certain affective states: for example, a beach scene or images of pets might be associated with happy emotions/positive affect, while arguments might be associated with anger/negative affect. Scene classifications were extracted using a ResNet50 model~\cite{he2016deep} trained on the Places 365 dataset~\cite{zhou2017places} while \textbf{Yolov5} \footnote{\url{https://github.com/ultralytics/yolov5}} was used for object detection. To obtain the action features, we first employed the popular open source shot detection method \textbf{SceneDetect} \footnote{\url{https://github.com/Breakthrough/PySceneDetect}} to decompose each music video into a sequence of scenes. Then, the I3D model described in~\cite{carreira2017quo} was used to compute the probability of each action class in each scene.

\section{Dataset analysis}\label{sec4}

\subsection{Distribution of emotion annotations}


In this section, we present high-level descriptive statistics for both the AV annotations and GEMS emotion labels. 

\subsubsection{Arousal-valence annotations}
\begin{figure}[ht]
\centering
    \includegraphics[scale=0.48]{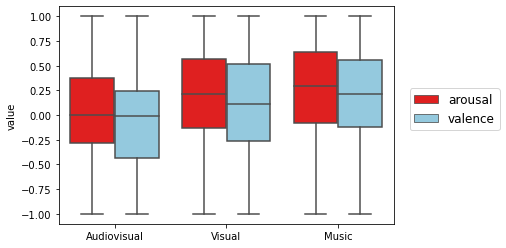}
    \caption{Distribution of arousal-valence annotations for each modality, sampled at 2Hz.}
    \label{fig:av_modality}
\end{figure}

The overall distribution of arousal-valence values across participants, sampled at 2Hz intervals, is displayed in Figure~\ref{fig:av_modality}. Across the entire dataset, the mean arousal value is 0.163 (music = 0.250, audiovisual = 0.192, visual= 0.044) while the mean valence is 0.073 (music = 0.177, audiovisual = 0.115, visual = -0.075). 

\subsubsection{GEMS emotion labels}
Across the entire dataset, participants selected an average of 4.06 labels per media stimulus (music = 3.92, audiovisual = 4.31, visual = 3.96). Each media stimulus was labeled with an average of 15.83 unique labels in total across participants (music = 15.36, audiovisual = 16.56, visual = 15.57). 

\begin{table}[ht!] \footnotesize
\centering \scriptsize
\begin{tabular}{@{}lccccc@{}}
\toprule
\textbf{GEMS labels} & \textbf{Music}   & \textbf{Audiovisual} & \textbf{Visual}  & \textbf{Total \%} & \multicolumn{1}{l}{\textbf{Songs \%}} \\ \midrule
\textbf{Wonder} &
  \multicolumn{1}{l}{} &
  \multicolumn{1}{l}{} &
  \multicolumn{1}{l}{} &
  \multicolumn{1}{l}{} &
  \multicolumn{1}{l}{} \\ 
\; Moved &
  54 &
  65 &
  63 &
  \cellcolor[HTML]{FFF8E2}3.03\% &
  12.33\% \\
\begin{tabular}[l]{@{}l@{}}\; Filled with \\ \; wonder\end{tabular} &
  49 &
  65 &
  68 &
  \cellcolor[HTML]{FFF8E2}3.03\% &
  12.33\% \\
\; Allured &
  50 &
  64 &
  55 &
  \cellcolor[HTML]{FFF8E2}2.82\% &
  11.45\% \\ \midrule
\textbf{Transcendence} &
  \multicolumn{1}{l}{} &
  \multicolumn{1}{l}{} &
  \multicolumn{1}{l}{} &
  \multicolumn{1}{l}{} &
  \multicolumn{1}{l}{} \\
\; Fascinated &
  79 &
  106 &
  71 &
  \cellcolor[HTML]{FFF8E2}4.27\% &
  17.34\% \\
\; Overwhelmed &
  59 &
  64 &
  52 &
  \cellcolor[HTML]{FFF8E2}2.92\% &
  11.86\% \\
\begin{tabular}[c]{@{}l@{}}\; Feeling of \\\; transcendence\end{tabular} &
  36 &
  39 &
  33 &
  \cellcolor[HTML]{FFF8E2}1.80\% &
  7.32\% \\ \midrule
\textbf{Peacefulness} &
  \multicolumn{1}{l}{} &
  \multicolumn{1}{l}{} &
  \multicolumn{1}{l}{} &
  \multicolumn{1}{l}{} &
  \multicolumn{1}{l}{} \\
\; Serene &
  36 &
  35 &
  33 &
  \cellcolor[HTML]{FFF8E2}1.73\% &
  7.05\% \\
\; Calm &
  54 &
  51 &
  53 &
  \cellcolor[HTML]{FFF8E2}2.63\% &
  10.70\% \\
\; Soothed &
  30 &
  33 &
  25 &
  \cellcolor[HTML]{FFF8E2}1.47\% &
  5.96\% \\ \midrule
\textbf{Tenderness} &
  \multicolumn{1}{l}{} &
  \multicolumn{1}{l}{} &
  \multicolumn{1}{l}{} &
  \multicolumn{1}{l}{} &
  \multicolumn{1}{l}{} \\
\; Tender &
  37 &
  60 &
  54 &
  \cellcolor[HTML]{FFF8E2}2.52\% &
  10.23\% \\
\; Affectionate &
  73 &
  104 &
  90 &
  \cellcolor[HTML]{FFF8E2}4.45\% &
  18.09\% \\
\; Mellow &
  41 &
  30 &
  44 &
  \cellcolor[HTML]{FFF8E2}1.92\% &
  7.79\% \\ \midrule
\textbf{Nostalgia} &
  \multicolumn{1}{l}{} &
  \multicolumn{1}{l}{} &
  \multicolumn{1}{l}{} &
  \multicolumn{1}{l}{} &
  \multicolumn{1}{l}{} \\
\; Nostalgic &
  105 &
  112 &
  101 &
  \cellcolor[HTML]{FFF8E2}5.30\% &
  21.54\% \\
\; Sentimental &
  79 &
  99 &
  87 &
  \cellcolor[HTML]{FFF8E2}4.42\% &
  17.95\% \\
\; Dreamy &
  85 &
  87 &
  74 &
  \cellcolor[HTML]{FFF8E2}4.10\% &
  16.67\% \\ \midrule
\textbf{Power} &
  \multicolumn{1}{l}{} &
  \multicolumn{1}{l}{} &
  \multicolumn{1}{l}{} &
  \multicolumn{1}{l}{} &
  \multicolumn{1}{l}{} \\
\; Strong &
  147 &
  164 &
  144 &
  \cellcolor[HTML]{FFF8E2}7.58\% &
  30.83\% \\
\; Energetic &
  193 &
  202 &
  178 &
  \cellcolor[HTML]{FFF8E2}9.55\% &
  38.82\% \\
\; Triumphant &
  48 &
  51 &
  52 &
  \cellcolor[HTML]{FFF8E2}2.52\% &
  10.23\% \\ \midrule
\textbf{Joyful Activation} & \multicolumn{1}{l}{} & \multicolumn{1}{l}{} & \multicolumn{1}{l}{} & \multicolumn{1}{l}{} & \multicolumn{1}{l}{}                  \\
\; Animated &
  43 &
  77 &
  57 &
  \cellcolor[HTML]{FFF8E2}2.95\% &
  11.99\% \\
\; Bouncy &
  113 &
  141 &
  105 &
  \cellcolor[HTML]{FFF8E2}5.98\% &
  24.32\% \\
\; Joyful &
  103 &
  106 &
  79 &
  \cellcolor[HTML]{FFF8E2}4.80\% &
  19.51\% \\ \midrule
\textbf{Sadness} &
  \multicolumn{1}{l}{} &
  \multicolumn{1}{l}{} &
  \multicolumn{1}{l}{} &
  \multicolumn{1}{l}{} &
  \multicolumn{1}{l}{} \\
\; Sad &
  96 &
  102 &
  98 &
  \cellcolor[HTML]{FFF8E2}4.93\% &
  20.05\% \\
\; Tearful &
  33 &
  35 &
  41 &
  \cellcolor[HTML]{FFF8E2}1.82\% &
  7.38\% \\
\; Blue &
  56 &
  46 &
  54 &
  \cellcolor[HTML]{FFF8E2}2.60\% &
  10.57\% \\ \midrule
\textbf{Tension} &
  \multicolumn{1}{l}{} &
  \multicolumn{1}{l}{} &
  \multicolumn{1}{l}{} &
  \multicolumn{1}{l}{} &
  \multicolumn{1}{l}{} \\
\; Tense &
  113 &
  110 &
  107 &
  \cellcolor[HTML]{FFF8E2}5.50\% &
  22.36\% \\
\; Agitated &
  59 &
  70 &
  57 &
  \cellcolor[HTML]{FFF8E2}3.10\% &
  12.60\% \\
\; Nervous &
  41 &
  49 &
  45 &
  \cellcolor[HTML]{FFF8E2}2.25\% &
  9.15\% \\ \midrule
\textbf{Total} &
  1,912 &
  2,167 &
  1,920 &
  5,999 &
  \multicolumn{1}{l}{} \\
\begin{tabular}[c]{@{}l@{}}\textbf{Average labels} \\\textbf{selected}\end{tabular} &
  3.92 &
  4.31 &
  3.96 &
  4.06 &
  \multicolumn{1}{l}{} \\ \bottomrule
\end{tabular}
\vspace{0.1cm}
\caption{Distribution of GEMS emotion labels for each modality.}
\label{t:emotion_label_correlation}
\end{table}

Across the dataset, the most frequently selected emotion labels were ``energetic'', which was selected for 38.82\% of the media stimuli and makes up 9.55\% of the labels, and ``strong'', which was selected for 30.83\% of the media stimuli and makes up 7.58\% of the labels. Both ``strong'' and ``energetic'' belong to the Power category, which was also the most frequently selected category with 55.3\% of discrete annotations containing at least one Power label. The least frequently selected labels were ``soothed'' and ``feeling of transcendence''.  The distribution of GEMS emotion labels selected might be a reflection of the media stimuli used, which all belonged to the pop music genre and are hence generally more upbeat.

\begin{figure}[ht!]
\centering
    \includegraphics[scale=0.34]{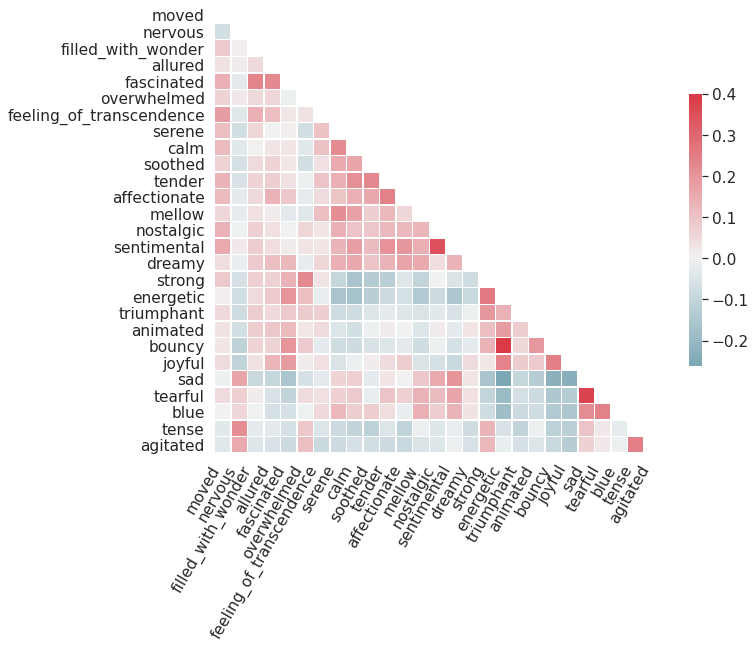}
    \caption{Pearson's correlations between categorical GEMS emotion labels.}
    \label{fig:emotion_label_correlation}
\end{figure}

We also computed the pairwise Pearson's correlation of the emotion labels selected by each participant for each stimulus. A heatmap of the results is displayed in Figure~\ref{fig:emotion_label_correlation}. The strongest positive correlations were between sad-tearful (r = 0.419, p \textless 0.001) and bouncy-energetic (r = 0.378, p \textless 0.001). One possible interpretation of the results is that these labels were frequently selected together. A second possibility is that the labels were selected by different people to indicate the same emotion for a particular media file, which might suggest partially redundant labels~\cite{aljanaki2016studying}. Indeed, the pair sad-tearful belongs to the same GEMS category ``Sadness'', while bouncy-energetic belongs to the same superfactor ``Vitality'' (encompassing Power and Joyful Activation), as shown in Table~\ref{t:discrete_emotions}. The strongest negative correlations were observed between energetic-sad (r = -0.255, p \textless 0.001) and joyful-sad (r = -0.206, p \textless 0.001), which is logical as the terms are antonyms semantically.

\subsection{Effect of annotation interface, media modality and annotator profile on perceived emotion}
We performed a series of analyses to examine whether the annotation interface, modality, and demographic and profile characteristics of participants influence the perceived emotion of media stimuli as measured by dynamic arousal-valence ratings as well as overall discrete emotion labels. 

\subsubsection{Arousal-valence annotations}
A visual inspection of the full time-series arousal-valence annotations reveals interesting differences between the average emotion trajectories in modalities. Although the annotations appear to be similar across all modalities for some stimuli, on the whole, annotations for the audiovisual modality tend to resemble those for the music modality. First, for each stimulus, we average the arousal annotations to obtain mean arousal ratings for the music, visual and audiovisual modalities. Then, we compute the Pearson's correlation between each pair of modalities. For arousal, the highest correlations were obtained for the music-audiovisual modality pair (mean = 0.731, sd = 0.309), followed by visual-audiovisual (mean = 0.479, sd = 0.453) then music-visual (mean = 0.411, sd = 0.436). We observe the same pattern for valence, where the highest correlations were again for the music-audiovisual modality pair (mean = 0.433, sd = 0.423), followed by visual-audiovisual (mean = 0.372, sd = 0.495) then music-visual (mean = 0.084, sd = 0.577).

Given that the arousal and valence annotations are time-series data, individual data points are not independent. As such, in order to perform the following set of statistical analyses, we pre-process all annotations by taking the median arousal and valence of each annotation sequence. A Kolmogorov-Smirnov test of goodness-of-fit indicates that median arousal (D = 0.273, p $<$ 0.001) and valence (D = 0.229, p $<$ 0.001) are not normally distributed. Hence, non-parametric statistical tests were used for the following analyses. 

\textbf{Overlay type:} Mann-Whitney U tests indicate that there is no significant difference between median arousal (U = $1.15 \times 10^5$, p = 0.2798) and valence (U = $1.14 \times 10^5$, p = 0.2456) for side-by-side versus overlaid annotation interfaces, which are relevant to the visual and audiovisual modalities. Hence, we combine the overlay types in our remaining analyses. The majority of participants (31 out of 49, $63.27\%$) indicated a preference for the annotation interface with the 2D valence-arousal axes located beside the video.

\textbf{Media modality:} The Kruskal-Wallis one-way ANOVA was run to compare arousal-valence annotations for stimuli presented in the music, visual, and audiovisual modalities. The results were significant for both arousal (H = 79.45, p $<$ 0.001) and valence (H = 100.38, p $<$ 0.001), indicating that there is a significant difference between median emotion annotations for at least two of the modalities. Arousal is highest in the music modality (median = 0.2957), followed by the audiovisual modality (median = 0.2087), and lowest in the visual modality (median = 0.0). Similarly, valence is highest in the music modality (mean = 0.2087), followed by the audiovisual modality (mean = 0.1130), and lowest in the visual modality (mean = -0.0087). This suggests that modality in which a media item is consumed exerts an effect on its perceived arousal and valence.

\subsubsection{GEMS emotion labels}
A contingency Chi-square test  was used to assess whether the frequency distribution of emotion labels is significantly different for media stimuli presented in different modalities. Additionally, we investigated whether extra musical factors (such as gender, musical training, and familiarity with the media item) and characteristics of the annotation interface (whether it was overlaid on the media stimulus or placed on the side) would influence the frequency of each emotion label's selection. We found that neither modality nor overlay type significantly influenced the frequency with which emotion labels were selected. However, the test statistic was significant for gender ($\chi^2 = 46.36$, df = 26, p = 0.008), formal musical training  ($\chi^2 = 93.43$, df = 26, p $<$ 0.001), and familiarity ($\chi^2 = 263.26$, df = 26, p $<$ 0.001).

\begin{figure*}[htb!]
\centering
    \includegraphics[width=\textwidth]{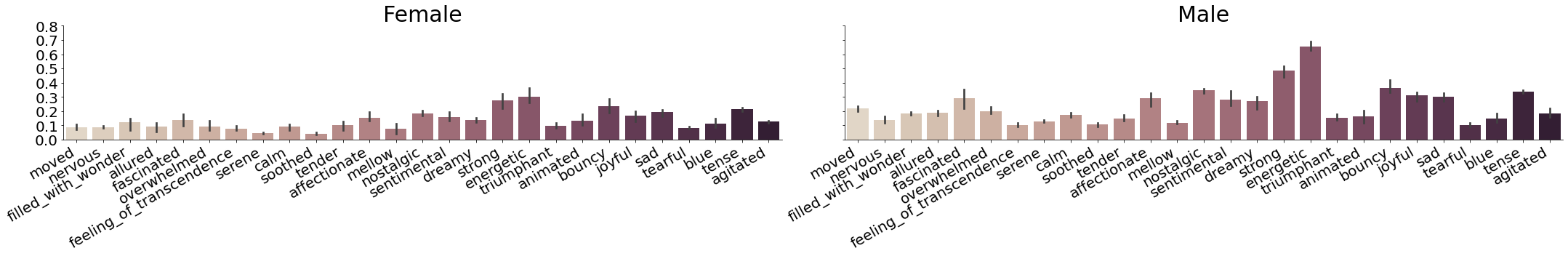}
    \caption{Distribution of GEMS emotion labels selected by female (left) and male (right) participants.}
    \label{fig:emotionLabelDist_gender}
\end{figure*}

We manually inspected the difference between expected and observed frequencies of emotion labels for each of these three groups and found that for gender (Figure~\ref{fig:emotionLabelDist_gender}), labels under the GEMS category Vitality such as ``energetic'', ``animated'' and ``bouncy'' tended to show the greatest deviation from the expected frequencies according to the Chi-square test. Interestingly, this difference was not consistent across gender: female participants selected the label ``energetic'' less frequently than male participants, while selecting ``animated'' and ``bouncy'' more frequently.

\begin{figure*}[htb!]
\centering
    \includegraphics[width=\textwidth]{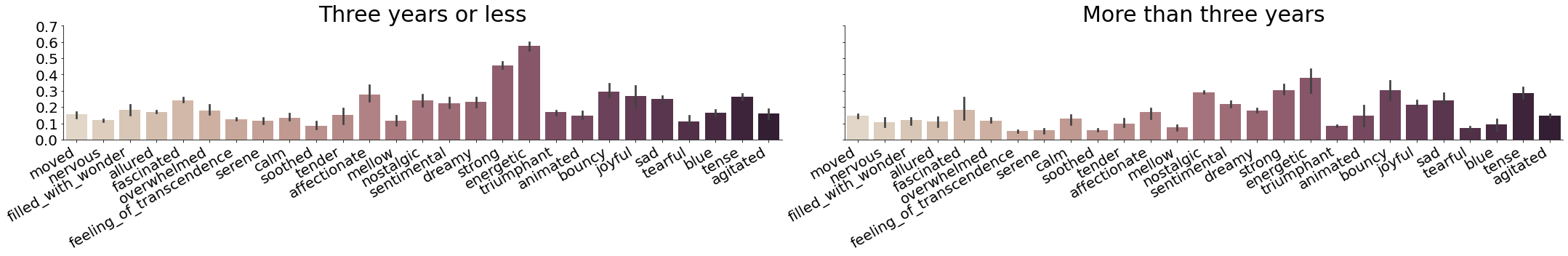}
    \caption{Left: Distribution of GEMS emotion labels selected by those with little (0 - 3 years) musical training; Right: more than 3 years of musical training.}
    \label{fig:emotionLabelDist_musicalTraining}
\end{figure*}

For `years of musical training' (Figure~\ref{fig:emotionLabelDist_musicalTraining}), the label ``nostalgic'' showed the greatest deviation from the expected frequencies according to the Chi-square test. Participants with three years or less of musical training selected the label ``nostalgic'' less frequently than expected, even though on average they were more likely to have seen or listened to the media stimuli previously.

\begin{figure*}[htb!]
\centering
    \includegraphics[width=\textwidth]{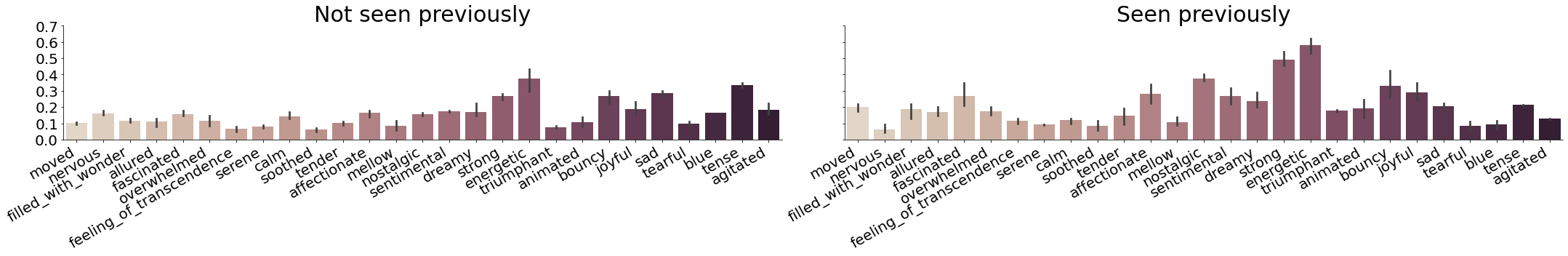}
    \caption{Left: Distribution of GEMS emotion labels selected by participant who were exposed to the media stimuli for the first time; Right: Participant has viewed or listened to the media item previously.}
    \label{fig:emotionLabelDist_familiarity}
\end{figure*}

For familiarity (Figure~\ref{fig:emotionLabelDist_familiarity}), the label ``tense'' shows the greatest deviation from the expected frequencies according to the Chi-square test. Participants who were unfamiliar with the media item selected labels from the GEMS category Tension such as ``tense'' and ``nervous'' much more frequently than expected, perhaps because they were less sure what to expect. They also selected the label ``nostalgic'' less frequently than expected, a result that is fairly intuitive since nostalgia is generally induced by familiarity. Finally, participants who were familiar with the media items selected labels from the GEMS category Sadness such as ``sad'' and ``blue'' more frequently than expected. 

\section{A predictive model augmented with isolated modality ratings (PAIR)}\label{sec5}
In this section, we present a novel approach for multimodal time-series emotion recognition, termed a Predictive model Augmented with Isolated modality Ratings (PAIR). We first discuss models for \textit{unimodal} emotion recognition relevant to the music and visual modalities, and culminate in a discussion of models for \textit{multimodal} emotion recognition relevant to the audiovisual modality. 

While sophisticated modelling techniques are undoubtedly required to improve state of the art performance on emotion recognition tasks, our focus here is to investigate whether an approach that utilizes parameters pre-trained on isolated modality ratings can lead to better performance on a multimodal (audiovisual) emotion recognition task. As such, all our models are based on long short-term memory (LSTM) networks~\cite{hochreiter1997long}. LSTMs are a popular variant of the recurrent neural network (RNN) \cite{williams1989learning} that can retain 'memory' of information seen previously over arbitrarily long intervals \cite{olah2015understanding}, and have been widely used to model time-series data for mood prediction and classification tasks (e.g.~\cite{weninger2014line,coutinho2015automatically, thao2019multimodal}). 

\subsection{Unimodal architectures}
Unimodal systems can be viewed as building blocks for a multimodal emotion recognition framework \cite{poria2017review}. In this case, we model arousal-valence annotations from the two unimodal conditions (music, visual) in MuVi. We term these annotations \textit{isolated modality ratings}, as they capture the emotion conveyed by the individual modalities that make up multimedia content in isolation. We trained separate models for arousal and valence in both isolated modalities. Isolated music modality ratings were predicted using audio features only while isolated visual modality ratings were predicted using visual features only, matching the information that human participants would have had access to during the experiment. 

\begin{figure*}[htb!]
\centering
    \includegraphics[scale=0.42]{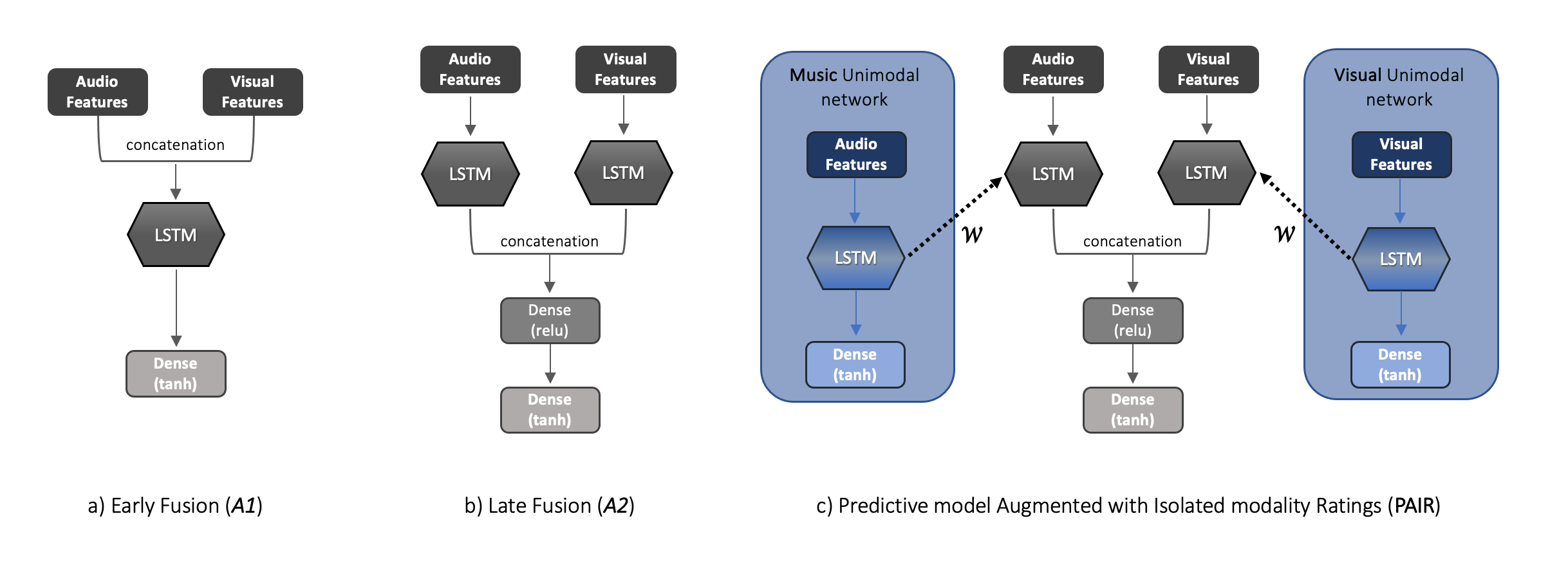}
    \caption{Illustrated workflow for the proposed Multimodal architectures for the audiovisual modality featuring: a) Early Fusion ($A1$), b) Late Fusion ($A2$), and c) the proposed PAIR. PAIR's LSTM blocks initializes the pre-trained weights ($w$) from the corresponding Music and Visual Unimodal networks.}
    \label{fig:PAIR_arch}
\end{figure*}

\subsection{Multimodal architectures}\label{sec5.2}

In this section, we model arousal-valence annotations from the multimodal (audiovisual) condition in MuVi, exhaustively considering all possible combinations of features. First, we examine model performance when using only audio features, and when using only visual features. Next, we compare the perfomance of three different model architectures for audio and visual feature fusion (denoted as \textbf{A1, A2 and PAIR}). A1 and A2 utilize early (feature-level) fusion and late (decision-level) fusion respectively, two of the most commonly-employed feature fusion methods \cite{poria2017review}. The proposed PAIR architecture builds on the unimodal models described above. Figure~\ref{fig:PAIR_arch}  presents the following architectures which are described in detail below:

\begin{itemize}
    \item \textbf{Early fusion (A1):} In this architecture, we apply an early fusion method by concatenating the audio and visual features to form a single input feature vector. The feature vector is then fed to the LSTM blocks. 
    \item \textbf{Late fusion (A2):} In this architecture,  we apply a late fusion method by feeding the two input modalities (i.e., audio and visual features) to separate LSTM blocks. Then, the outputs of both LSTMs are concatenated and fed to a final, fully-connected layer for prediction.
    \item \textbf{PAIR:} This architecture is identical to $A2$, except that the LSTM blocks are initialized with pre-trained weights from the corresponding unimodal models (trained on ratings from the isolated music and visual modalities). Then, the weights are fine-tuned on ratings from the multimodal audiovisual modality. The motivation for this architecture is twofold: firstly, from a computational perspective, the amount of data in MuVi is relatively small for training deep learning models. Additionally, there is evidence from neuroimaging studies that the integration of emotional information from auditory and visual channels can be attributed to the interaction between unimodal auditory and visual cortices plus higher order supramodal cortices, rather than from direct crosstalk between the auditory and visual cortices \cite{kreifelts2007audiovisual}. As such, we hypothesized that we may obtain better performance on a multimodal affect recognition task by mimicking this neural architecture and initializing a multimodal model using weights pre-trained on relatively 'pure' isolated modality ratings.
\end{itemize}

\subsection{Implementation details}
In all architectures (i.e., unimodal and multimodal) an \textbf{LSTM block} consists of two stacked LSTM layers each of size 256, with dropout of 0.2 to prevent overfitting. The output of the LSTM layers is fed to a fully connected layer with a ``tanh'' activation that outputs the final prediction. This value lies in the range of $[-1, 1]$, matching that of the arousal and valence annotations. In the Late Fusion (A2) and PAIR (A3) architectures, we have an additional fully connected layer before the final predictive layer with 256 hidden units and ``relu'' activation.

During the optimisation phase, the Adam~\cite{kingma2014adam} optimizer was used with a learning rate of 0.0001, while ``Mean Squared Error'' (MSE) was used as the loss function. Since the annotation and the feature extraction process are both dynamic, based on timesteps of 0.5s, our benchmark models predict arousal and valence at the next time step by taking the information of previous time-steps as input (i.e., sequence length of the LSTM). After experimenting, we found that the optimal sequence length for both arousal and valence across all modalities is 4 timesteps (i.e., 2 seconds). The model was implemented using the Tensorflow 2.x~\cite{abadi2016tensorflow} deep learning framework, and is available online \footnote{\url{https://github.com/AMAAI-Lab/MuVi}}.

\section{Experimental Setup}\label{sec6}
In this section, we introduce our experimental design to evaluate our predictive models and examine the effectiveness of the novel transfer-learning that PAIR architecture utilizes. Moreover, we introduce the equivalent linear models as baselines, not only for comparison with the proposed PAIR architecture, but also for finding the most influential features.

\subsection{Baseline: Linear Models}

We implemented linear LASSO regression models as a simple baseline predictive model for valence and arousal. In addition to investigating the relative importance of audio and visual features for predicting emotion in audiovisual media, we are also interested in examining the individual features that are most predictive of arousal and valence across modalities. 

In a review of music emotion recognition methods, \cite{yang2018review} found that the use of too many features generally results in performance degradation. Feature selection methods such as manual selection or removal of highly correlated features can help reduce computational complexity and improve model performance by mitigating overfitting issues. Additionally, all possible combinations of modalities and corresponding feature sets are examined.
We use the least absolute shrinkage and selection operator (LASSO) as our baseline linear predictive model, as it can also be used for feature selection~\cite{muthukrishnan2016lasso}. The LASSO minimizes the absolute sum of all coefficients (L1 regularization), and if subsets of features are highly correlated, the model tends to ‘select' one feature from the pool while shrinking the remaining coefficients to zero. Cross-validation was performed to determine optimal values for the complexity parameter alpha, which controls the strength of the regularization.

\begin{table*}[htb!]
\centering
\begin{tabular}{@{}
>{\columncolor[HTML]{FFFFFF}}c@{\hskip 0.45in} 
>{\columncolor[HTML]{FFFFFF}}c@{\hskip 0.45in}  
>{\columncolor[HTML]{FFFFFF}}c@{\hskip 0.45in} 
>{\columncolor[HTML]{FFFFFF}}c@{\hskip 0.45in} 
>{\columncolor[HTML]{FFFFFF}}c@{\hskip 0.45in} 
>{\columncolor[HTML]{FFFFFF}}c @{}}
\toprule
\cellcolor[HTML]{FFFFFF}{\color[HTML]{000000} } &
  \cellcolor[HTML]{FFFFFF}{\color[HTML]{000000} } &
  \multicolumn{2}{c}{\cellcolor[HTML]{FFFFFF}{\color[HTML]{000000} \textbf{Arousal}}} &
  \multicolumn{2}{c}{\cellcolor[HTML]{FFFFFF}{\color[HTML]{000000} \textbf{Valence}}} \\ \cmidrule(l){3-6} 
\multirow{-2}{*}{\cellcolor[HTML]{FFFFFF}{\color[HTML]{000000} \textbf{Modality}}} &
  \multirow{-2}{*}{\cellcolor[HTML]{FFFFFF}{\color[HTML]{000000} \textbf{Features}}} &
  {\color[HTML]{000000} RMSE} &
  {\color[HTML]{000000} CCC} &
  {\color[HTML]{000000} RMSE} &
  {\color[HTML]{000000} CCC} \\ \midrule
{\color[HTML]{000000} Music} &
  {\color[HTML]{000000} Audio} &
  {\color[HTML]{000000} 0.2675 $\pm$ 0.1075} &
  {\color[HTML]{000000} 0.3417 $\pm$ 0.2909} &
  {\color[HTML]{000000} 0.1573 $\pm$ 0.1029} &
  {\color[HTML]{000000} 0.1196 $\pm$ 0.1950} \\
{\color[HTML]{000000} Visual} &
  {\color[HTML]{000000} Visual} &
  {\color[HTML]{000000} 0.4110 $\pm$ 0.1571} &
  {\color[HTML]{000000} 0.0785 $\pm$ 0.1579} &
  {\color[HTML]{000000} 0.3506 $\pm$ 0.1283} &
  {\color[HTML]{000000} 0.0415 $\pm$ 0.1332} \\
{\color[HTML]{000000} Audiovisual} &
  {\color[HTML]{000000} Audio} &
  {\color[HTML]{000000} 0.2721 $\pm$ 0.1145} &
  {\color[HTML]{000000} 0.3095 $\pm$ 0.3218} &
  {\color[HTML]{000000} 0.4106 $\pm$ 0.1330} &
  {\color[HTML]{000000} 0.0260 $\pm$ 0.1419} \\
{\color[HTML]{000000} Audiovisual} &
  {\color[HTML]{000000} Visual} &
  {\color[HTML]{000000} 0.3649 $\pm$ 0.1744} &
  {\color[HTML]{000000} 0.0772 $\pm$ 0.1475} &
  {\color[HTML]{000000} 0.3834 $\pm$ 0.1427} &
  {\color[HTML]{000000} 0.0438 $\pm$ 0.1210} \\
{\color[HTML]{000000} Audiovisual} &
  {\color[HTML]{000000} Audio and Visual} &
  {\color[HTML]{000000} 0.2783 $\pm$ 0.1231} &
  {\color[HTML]{000000} 0.3070 $\pm$ 0.3127} &
  {\color[HTML]{000000} 0.3831 $\pm$ 0.1417} &
  {\color[HTML]{000000} 0.0503 $\pm$ 0.1680} \\ \bottomrule
\end{tabular}

\vspace{0.1cm}
\caption{Evaluation of the LASSO linear models in terms of RMSE and CCC (mean $\pm$ standard deviation). Note that lower values of RMSE, and higher values of CCC, reflect higher predictive accuracy.}
\label{t:linear_results}
\end{table*}

\begin{figure*}[htb!]
\centering
    \includegraphics[width=\textwidth]{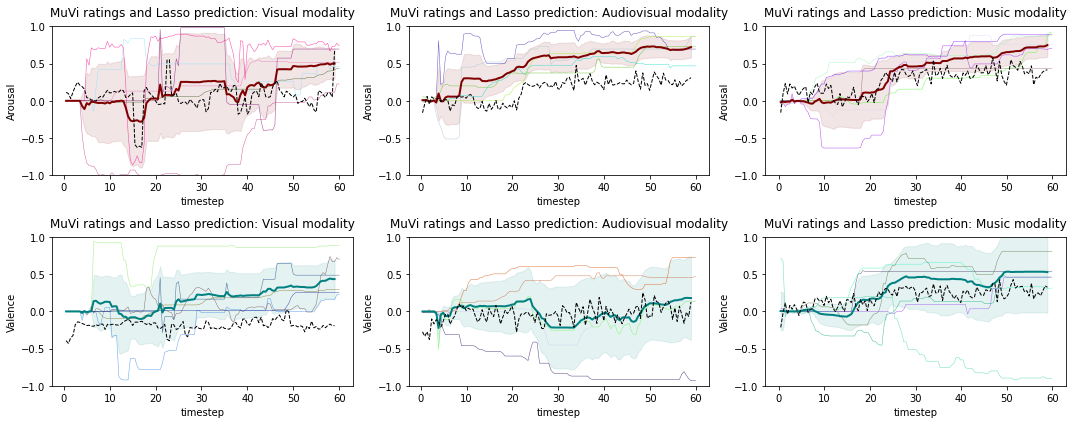}
    \caption{LASSO model predictions of arousal and valence ratings in the visual (left), audiovisual (center) and music (right) modalities for the media item 'Living On A Prayer' by Bon Jovi. Each colored line in the background represents an individual participants' rating, while the bold red (arousal) and teal (valence) lines represent the Evaluator Weighted Estimator rating along with its standard deviation. The dashed black lines represent the model's predicted rating at that timestep.}
    \label{fig:av_sample_lasso}
\end{figure*}

\subsection{Predictive Model Evaluation Metrics}
To account for variance in the media items and fully utilize the information provided by the dataset, the PAIR and linear models are evaluated using k-fold cross-validation. We partitioned the media items into five folds using the KFold iterator in sklearn \footnote{\url{https://scikit-learn.org/stable/modules/cross_validation.html#k-fold}} with random state set to 42, iteratively holding out each fold as a validation set for the model trained on the remaining four fold.

Regarding metrics, we modified the approach taken by~\cite{aljanaki2017developing} to design benchmarks for the MediaEval Emotion in Music task, and we use two evaluation metrics to benchmark the performance of baseline methods: root mean square error (RMSE) and Lin's Concordance Correlation Coefficient (CCC). Intuitively, the RMSE is an indicator of how well the predicted emotion matches the gold-standard or ``true'' emotion of a media item, while the CCC (inter-rater agreement, or the extent to which the human raters provide similar ratings across trials) was initially designed as a reproducibility index and quantifies the agreement between two vectors by measuring their difference from the 45 degree line through the origin. As such, it captures both the pattern of association and systematic differences between two sets of points. The CCC for two vectors $(r_1, r_2)$ is calculated as:

\begin{equation}
CCC (r_1,r_2)= \frac{2 Corr(r_1, r_2)\sigma_{r_1}\sigma_{r_2}}{\sigma^2_{r_1} + \sigma^2_{r_2} + (\mu_{r_1} - \mu_{r_2})^2}
\end{equation}

RMSE and predicted CCC are computed for each media item in each validation set. The mean and standard deviations of each evaluation metric are then computed across all cross-validation folds.

A gold-standard annotation sequence to be used as the modelling target was computed for each media item in each modality using the Evaluator Weighted Estimator (EWE) approach \cite{grimm2007primitives}. As its name suggests, the EWE is a weighted average which gives more reliable annotations a higher weight. The weight of an annotation $r_j$ is calculated as the correlation between $r_j$ and the unweighted average of all annotations $\bar{r}$. It is plausible that the dataset contains unreliable annotations with $w_j<0$; in the following analyses we deal with unreliable annotations by setting their weight to zero such that they are effectively not taken into account when calculating the gold-standard annotation sequence. 

\begin{equation}
w_j = Corr(r_j, \bar{r})
\end{equation}

\begin{equation}
r_{EWE} = \frac{1}{\sum_j w_j} \sum_j w_j r_j
\end{equation}

\subsection{Human inter-rater agreement}

\begin{table}[htb!]
\centering
\begin{tabularx}{\linewidth}{@{}l@{\hskip 0.77in}c@{\hskip 0.77in}c@{}}
\toprule
Modality    & Arousal & Valence \\ \midrule
Music       & 0.4062 $\pm$0.32 & 0.2385 $\pm$0.31 \\
Visual & 0.2839 $\pm$0.32 & 0.3115 $\pm$0.32 \\
Audiovisual       & 0.3369 $\pm$0.32 & 0.2384 $\pm$0.32 \\ \bottomrule
\end{tabularx}%

\vspace{0.1cm}
\caption{Evaluation of inter-rater agreement by calculating the CCC of each annotation for all modalities.}
\label{t:inter_rater_eval}
\end{table}

We also calculate inter-rater agreement, which provides a useful auxiliary metric for comparison with the predictive model results. Inter-rater agreement indicates the extent to which the perception of emotional content of a media item is similar between human raters, and may serve as a measure of the quality of the human annotations of the dataset. We calculate the inter-rater agreement as $CCC(r_j, r_{EWE})$. The CCC of each annotation $r_j$ with the corresponding $EWE$ for the media stimuli whereby the EWE was calculated without $r_j$ so as to avoid overstating inter-rater agreement. Agreement was low to moderate. Across the entire dataset, the mean CCC for arousal was 0.34 $\pm$ 0.32 and the mean CCC for valence was 0.26 $\pm$ 0.32.

The results in Table~\ref{t:inter_rater_eval} are in line with previous empirical studies~\cite{yang2008regression} which show that raters tend to demonstrate greater agreement in their perceptions of arousal in music as compared to valence, possibly because auditory cues relevant to arousal such as tempo and pitch are more salient and the perception of arousal is hence less subjective than that of valence. There was also greater agreement for arousal compared to valence in the audiovisual condition. 


\section{Results and Discussion}\label{sec7}

\subsection{Predicting valence and arousal}
\subsubsection{Baseline: LASSO models}

Starting with the baselines, Table~\ref{t:linear_results} summarises the results of the LASSO models in different modality configurations. As expected, the best model performance was observed for arousal in the music modality. A t-test indicates that the simple linear model produced results that were not significantly different from human-level performance as determined by annotator's inter-rater agreement ($t=1.82, p > 0.05$). Similarly, linear model performance was not significantly different from human-level performance when predicting arousal in the video modality with audio features only, and with both audio and visual features. However, none of the linear models were able to predict valence accurately enough to match human-level performance. These results agree with existing empirical work on music emotion recognition that typically achieves much higher accuracy for arousal compared to valence~\cite{ehrlich2019closed}. 

A visual inspection of the linear model predictions in Figure \ref{fig:av_sample_lasso} reveals that even though the predicted annotations are not significantly different from those of human raters, they are unrealistic looking and much less smooth, possibly because the model does not account for features at the preceding timesteps.


\begin{table*}[]
\centering
\begin{tabular}{@{}
>{\columncolor[HTML]{FFFFFF}}c@{\hskip 0.45in} 
>{\columncolor[HTML]{FFFFFF}}c@{\hskip 0.45in}  
>{\columncolor[HTML]{FFFFFF}}c@{\hskip 0.45in} 
>{\columncolor[HTML]{FFFFFF}}c@{\hskip 0.45in} 
>{\columncolor[HTML]{FFFFFF}}c@{\hskip 0.45in} 
>{\columncolor[HTML]{FFFFFF}}c @{}}
\toprule
\cellcolor[HTML]{FFFFFF}{\color[HTML]{000000} } &
  \cellcolor[HTML]{FFFFFF}{\color[HTML]{000000} } &
  \multicolumn{2}{c}{\cellcolor[HTML]{FFFFFF}{\color[HTML]{000000} \textbf{Arousal}}} &
  \multicolumn{2}{c}{\cellcolor[HTML]{FFFFFF}{\color[HTML]{000000} \textbf{Valence}}} \\ \cmidrule(l){3-6} 
\multirow{-2}{*}{\cellcolor[HTML]{FFFFFF}{\color[HTML]{000000} \textbf{Modality}}} &
  \multirow{-2}{*}{\cellcolor[HTML]{FFFFFF}{\color[HTML]{000000} \textbf{Features}}} &
  {\color[HTML]{000000} RMSE} &
  {\color[HTML]{000000} CCC} &
  {\color[HTML]{000000} RMSE} &
  {\color[HTML]{000000} CCC} \\ \midrule
{\color[HTML]{222222} Music} &
  Audio &
  \textbf{0.0973} $\pm$ \textbf{0.0814} &
  0.4080 $\pm$ 0.2631 &
  \textbf{0.1331} $\pm$ \textbf{0.1225} &
  \textbf{0.1741} $\pm$ \textbf{0.2142} \\
Visual &
  Visual &
  0.1625 $\pm$ 0.0853 &
  0.1417 $\pm$ 0.2075 &
  0.2340 $\pm$ 0.1341 &
  0.0675 $\pm$ 0.2294 \\
Audiovisual &
  Audio &
  0.1109 $\pm$ 0.0853 &
  0.3715 $\pm$ 0.2455 &
  0.1970 $\pm$ 0.1264 &
  0.0497 $\pm$ 0.1523 \\
Audiovisual &
  Visual &
  0.2434 $\pm$ 0.1513 &
  0.0827 $\pm$ 0.2231 &
  0.2767 $\pm$ 0.1493 &
  0.0449 $\pm$ 0.1859 \\ \midrule
Audiovisual &
  Audio and Visual (A1) &
  0.1285 $\pm$ 0.0969 &
  0.3177 $\pm$ 0.2824 &
  0.2264 $\pm$ 0.1347 &
  0.0734 $\pm$ 0.2136 \\ 
Audiovisual &
  Audio and Visual (A2) &
  0.1372 $\pm$ 0.1041 &
  0.3230 $\pm$ 0.2882 &
  0.2392 $\pm$ 0.1390 &
  0.0708 $\pm$ 0.2191 \\
Audiovisual  &
  Audio and Visual (PAIR) &
  0.1075 $\pm$ 0.0847 &
  \textbf{0.4102} $\pm$ \textbf{0.2488} &
  0.2126 $\pm$ 0.1268 &
  0.0886 $\pm$ 0.2198 \\ \bottomrule
\end{tabular}%
\vspace{0.1cm}
\caption{Evaluation of the LSTM models (unimodal and mulitimodal) in terms of RMSE and CCC (mean $\pm$ standard deviation) featuring A1, A2 and the novel PAIR architectures in the audiovisual modality.}
\label{t:lstm_results}
\end{table*}

\subsubsection{PAIR (LSTM) models}
\begin{figure}[ht]
\centering
    \includegraphics[scale=0.65]{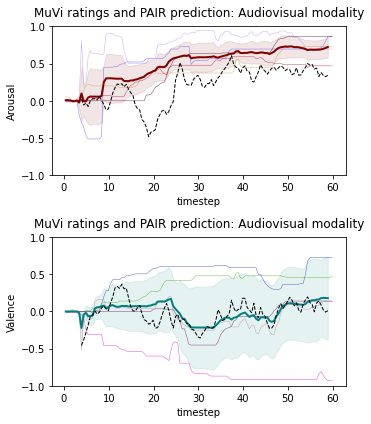}
    \caption{PAIR model predictions of arousal and valence ratings in the audiovisual modality for Bon Jovi's Living On A Prayer. Similarly to \ref{fig:av_sample_lasso}, each colored line in the background represents an individual participants' rating, while the bold red (arousal) and teal (valence) lines represent the Evaluator Weighted Estimator rating along with its standard deviation. The dashed black lines represent the model's predicted rating at that timestep.}
    \label{fig:av_sample_pair}
\end{figure}
Table~\ref{t:lstm_results} shows the results of the proposed LSTM models for all modality-feature combinations examined using the linear models, and for the three additional architectures (A1, A2, PAIR) described in Section~\ref{sec5.2}. Generally, the LSTM models improve over the linear model’s results in all modality-feature combinations. As with the linear models, the best performance was observed when predicting arousal in the music modality, and predictions of arousal were more accurate than predictions of valence. The worst performance was observed in the visual modality in which only visual features were available (i.e., a unimodal architecture).

Next, we compare the A1 (early fusion), A2 (late fusion) and PAIR architectures for multimodal emotion recognition. For both arousal and valence, the early and late fusion architectures improve over the linear model’s results in terms of RMSE, but not for CCC. However, the PAIR architecture initialized with weights pre-trained on the unimodal LSTM networks produces a noticeable improvement in arousal prediction in terms of CCC. Paired t-tests show that the PAIR is significantly different (better) than the linear models for predicting arousal in terms of CCC ($t(40) = 2.81, p<0.01$) and RMSE ($t(40)=-13.0, p<.001$), and for predicting valence in terms of CCC ($t(40) = -11.7, p<0.001$), highlighting the effectiveness of our proposed transfer learning approach leveraging isolated modality ratings. Similarly, the PAIR model predictions in Figure \ref{fig:av_sample_pair} appear to be less erratic than the linear model predictions, and more closely follows the contours of the human participants' ratings.

\subsection{Most influential features}

\begin{table}[]
\centering
\begin{tabular}{@{}lccc@{}} \toprule
\textbf{Modality} &
  \textbf{Features} &
  \textbf{Arousal} &
  \textbf{Valence} \\ \midrule
Music &
  Audio &
  \begin{tabular}[c]{@{}c@{}}MFCC, loudness, \\ line spectral \\ frequencies\end{tabular} &
  \begin{tabular}[c]{@{}c@{}}Line spectral \\ frequencies, \\ loudness, MFCC\end{tabular} \\ \midrule
\begin{tabular}[c]{@{}l@{}}Visual\end{tabular} &
  Visual &
  \begin{tabular}[c]{@{}c@{}}Actions, lighting,\\  scene\end{tabular} &
  \begin{tabular}[c]{@{}c@{}}Lighting, scene, \\ hue and saturation, \\ objects, actions\end{tabular} \\ \midrule
Audiovisual &
  Audio &
  \begin{tabular}[c]{@{}c@{}}MFCC, loudness, \\ F0, line spectral \\ frequencies\end{tabular} &
  \begin{tabular}[c]{@{}c@{}}Intensity, MFCC, \\ line spectral \\ frequencies, \\ zero crossing rate\end{tabular} \\ \midrule
Audiovisual &
  Visual &
  \begin{tabular}[c]{@{}c@{}}Actions, hue, \\ lighting, scene\end{tabular} &
  \begin{tabular}[c]{@{}c@{}}Actions, lighting, \\ hue and saturation, \\ scene\end{tabular} \\ \midrule
Audiovisual &
  \begin{tabular}[c]{@{}c@{}}Audio \& \\ Visual\end{tabular} &
  \begin{tabular}[c]{@{}c@{}}MFCC, loudness, \\ actions, line spectral\\ frequencies, F0\end{tabular} &
  \begin{tabular}[c]{@{}c@{}}Lighting, actions,\\ hue and saturation, \\ intensity \end{tabular} \\ \bottomrule
\end{tabular} 
\vspace{0.1cm}
\caption{Most predictive features for Valence and Arousal according to linear models.}
\label{t:predictive_fes}
\end{table}

LASSO performs feature selection by shrinking the coefficients of variables that are less significant to zero~\cite{muthukrishnan2016lasso}. We report the features with the largest absolute coefficient values for each model in Table~\ref{t:predictive_fes}. Coefficient values were calculated by averaging across all cross-validation folds.

When annotators were presented with audiovisual stimuli, the variance in arousal values seems to be explained largely by audio features while the variance in valence can be largely expained by visual features. This is supported by the observation that models trained on target labels from the audiovisual modality with audio features only outperformed those trained with visual features only when predicting arousal, but not when predicting valence.

Across the media modalities that included visual features, the LASSO regression models consistently estimated coefficients of relatively large magnitude for several action classes extracted using the I3D model \cite{carreira2017quo} including ``stretching arm'' and ``sneezing''. We manually inspected the video scenes where these actions were classified as highly likely. The sign of the coefficient for ``stretching arm'' is negative, indicating that the action is associated with decreased arousal and valence. In the video scenes, ``stretching arm'' seemed to correspond generally to moving of the arms in the context of actions such as embraces or dancing. Additionally, most of the scenes were shot at least partially in slow motion. Previous work on affective gestures~\cite{glowinski2011toward} has demonstrated that the spatial extension of movements~\cite{wallbott1998bodily} and openness of arm arrangements~\cite{mehrabian2017nonverbal} express emotionally relevant information by indicating attributes of the communicator such as accessibility. The tempo conveyed through cinematic techniques such as camera and subject movement has also been found to be a highly expressive aspect of videos~\cite{wang2015video}. It might be possible that it is not the action of stretching one's arm that conveys affective information, but the action classification model is recognizing latent dimensions such as open gestures and slowed motion that convey anticipation or affective states such as relaxedness that correspond to lowered arousal and valence. 

The sign of the coefficient for ``sneezing'' is also negative. In the video scenes, ``sneezing'' seemed to correspond generally to shots centering on a single actor in which facial expressions were clearly visible. One possible reason why the feature could be associated with lower arousal and valence is because the actors depicted were usually trying to express seriousness or intensity. The reader may ask: if this were the case, why was the model assigning much higher weights to facial actions than facial expression features, which were also available? 
Action features were extracted from scenes that could stretch over longer durations, while facial expressions were extracted from still frames sampled from the underlying video at regular intervals. Consequently, predicted facial expressions could change significantly between consecutive sampled frames while predicted action features tended to be more stable. The stability of the action features might hence be a better match to the human annotations of arousal and valence which tend to change smoothly over time. This finding further justifies the use of time-series models that can incorporate information provided in previous time-steps for emotion recognition.

\section{Conclusion}\label{sec8}
In this paper, we explore the relative contribution of audio and visual information on perceived emotion in media items. To do so, we provide a new dataset of music videos labeled with both continuous as well as discrete measures of emotion in both isolated and combined modalities. We then used the dataset to train two types of emotion recognition models: interpretable linear predictive models for feature importance, and more sophisticated LSTM models. We trained three LSTM models for audiovisual emotion recognition: early fusion ($A1$), late fusion ($A2$), and a novel transfer learning model pre-trained on isolated modality ratings which we named \textbf{PAIR}. Notably, we find that transfer effects from isolated modalities can enhance performance of a multimodal model through transfer learning, a result that could not be shown from previous datasets due to the lack of isolated modality ratings. 

Our analyses show that affective multimedia content analysis is a complex problem for several reasons, including the subjectivity of perceived emotion, as evidenced by low to moderate agreement between human annotators using both continuous and discrete measures of emotion \cite{ong2019modeling}\cite{aljanaki2017developing}, and the association between factors such as gender and familiarity with perceived emotion \cite{yang2007music}. Specifically, we find that gender, familiarity, and musical training have a significant influence on the emotion labels selected by participants. In the current study, we randomly assigned stimuli to participants and focused on modeling the average emotion ratings reported. However, future work could examine methods to integrate profile and demographic information in addition to data-based features to build personalized models, through model adaptation \cite{wang2012personalized} or other techniques. 

Our work investigated several visual context features rarely used in affect detection including scene, object and action features. We find evidence that action features are an important source of affective information, and that scene-based features might be more appropriate for time-series emotion recognition than features extracted from windows of fixed duration. These results add to a growing body of work showing that people make use of contextual information when making affective judgements \cite{kosti2019context}. 

Further, although the visual modality is generally considered to be dominant ~\cite{posner1976visual}~\cite{tsay2013sight}, we find that the auditory modality explains most of the variance in arousal for multimedia stimuli, while both the auditory and visual modalities contribute to explaining the variance in valence. Future work should investigate the impact of incongruent information across modalities, and explore which features cause the user's responses to be more oriented towards the visual or auditory modality when multimedia stimuli are presented, as this may may clarify when visual or auditory dominance occurs. A model that is able to fully capture the influence of emotion of each modality may even be used as a conditional generative model for music~\cite{herremans2017morpheus, tan2020music, makris2021generating} or video. Additionally, although we have begun to examine feature importance, there is still room for future work on feature engineering. For example, extracting more extensive deep pretrained repesentations of latent audio and video dimensions might enable models to take into account more complex information. 

The isolated modality ratings revealed several interesting patterns. Compared to listening to music alone, we find that the addition of a visual modality in music videos generally produces slightly lower arousal and valence and lower interrater agreement for arousal. In line with previous work \cite{yang2012machine}, agreement for valence was lower than agreement for arousal in the music and audiovisual modalities. In the visual modality, however, agreement for valence was higher than agreement for arousal. In fact, the visual modality produced the lowest agreement in arousal but the highest agreement in valence. Taken together with the feature importance results above, the data suggests that visual information contributes to people’s perceptions of valence more than arousal. 

Our results also have several practical implications. Convergent evidence across multiple analyses strongly suggests that in music videos, our perception of arousal is largely determined by the music while both the music and visual narrative contribute to our perception of valence. Additionally, an individual’s impression of the emotion conveyed by a song could be very different depending on the modality in which it is consumed. Content platforms may find it useful to incorporate modality-related information into their search and recommendation algorithms. Similarly, music video producers might benefit from a greater awareness of each modality’s contribution to the emotion conveyed by their content, and to the different modalities in which their content can be consumed. Of course, music videos are a subset of audiovisual content and, more broadly, multimodal content \cite{poria2017review}. While music videos usually involve the music being produced first with the visual narrative subsequently designed around the song, other scenarios are also possible during audiovisual content production: the video may be shot first, with music selection occurring subsequently, or the video and music production may proceed in parallel and inform one another.  Investigating isolated modality ratings in other types of audiovisual and multimodal content is an important avenue for future work. We hope that our research and dataset will provide a useful steppdfing-stone towards advancing the field of affective computing through the study of isolated modalities.

\ifCLASSOPTIONcompsoc
  \section*{Acknowledgments}
\else
  \section*{Acknowledgment}
\fi

This work was supported by the RIE2020 Advanced Manufacturing and Engineering (AME) Programmatic Fund (No. A20G8b0102), Singapore, as well as the Singapore Ministry of Education, Grant no. MOE2018-T2-2-161.

\ifCLASSOPTIONcaptionsoff
  \newpage
\fi



\bibliographystyle{IEEEtran}
\bibliography{main.bib}
%

%

\begin{IEEEbiography}[{\includegraphics[width=1in,height=1.25in,clip,keepaspectratio]{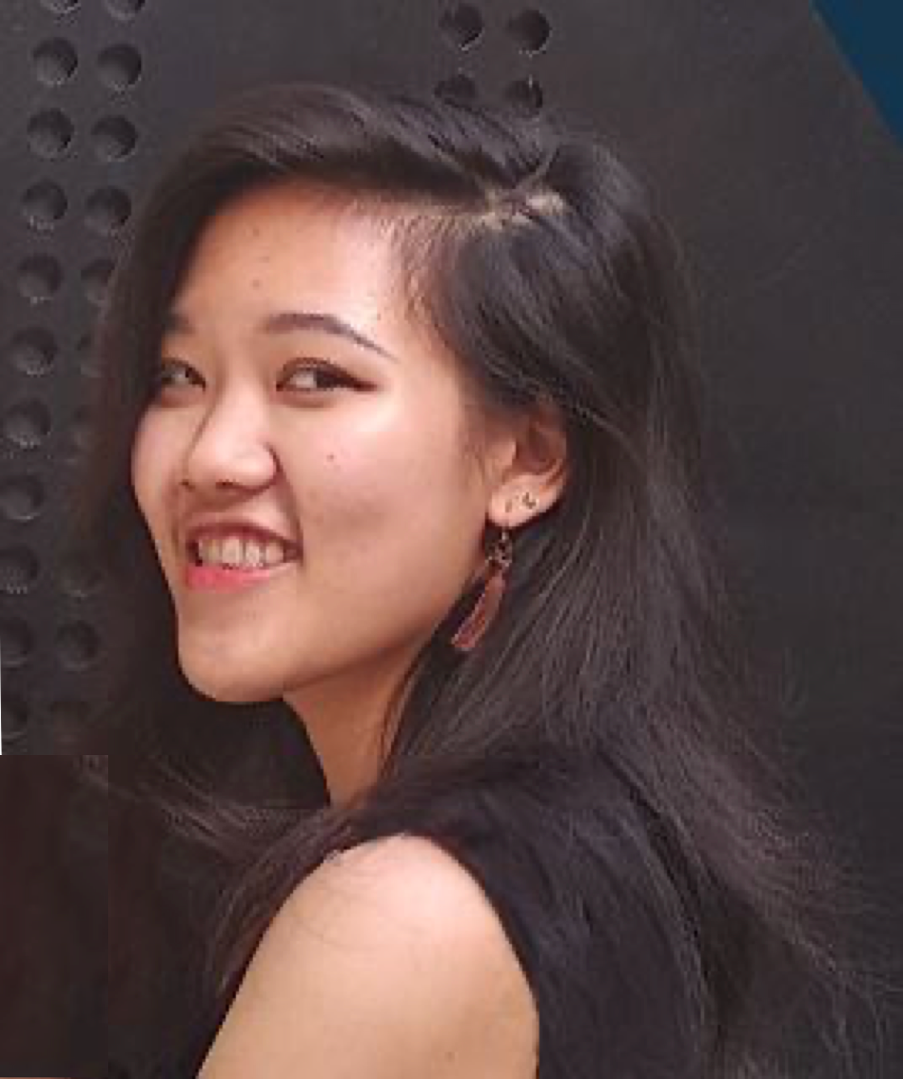}}]{Phoebe Chua} received her M.Sc. in Information Systems and Analytics at the National University of Singapore, and is currently pursuing a PhD in Information Systems and Analytics with a focus on computational social science at the National University of Singapore under the mentorship of Professor Desmond C. Ong. Her research interests include understanding emotion in the contexts of interpersonal relationships and aesthetic experiences. 
\end{IEEEbiography}

\begin{IEEEbiography}[{\includegraphics[width=1in,height=1.25in,clip,keepaspectratio]{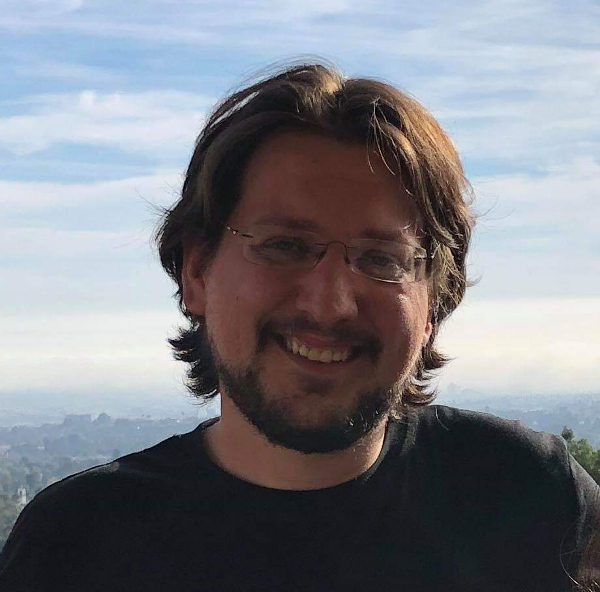}}]{Dr. Dimos (Dimosthenis) Makris} is an active researcher in the field of Music Information Retrieval. His PhD research includes A.I. applications for music generation using symbolic data, dataset creation, and track separation/instrument recognition tasks. After his PhD, he worked as a postdoctoral researcher at the Singapore University of Technology and Design under the supervision of Assistant Professor Dorien Herremans. He also has experience in the music industry as a Technical Director of Mercury Orbit Music, an A.I. music generation start-up company (2017-19, 2021-now), and as a Recording Engineer/Producer for over seven years. His current research interests include Deep Learning architectures for conditional music generation and exploring efficient encoding representations. Finally, he holds a diploma in Music Theory and is an active piano player.
\end{IEEEbiography}

\begin{IEEEbiography}[{\includegraphics[width=1in,height=1.25in,clip,keepaspectratio]{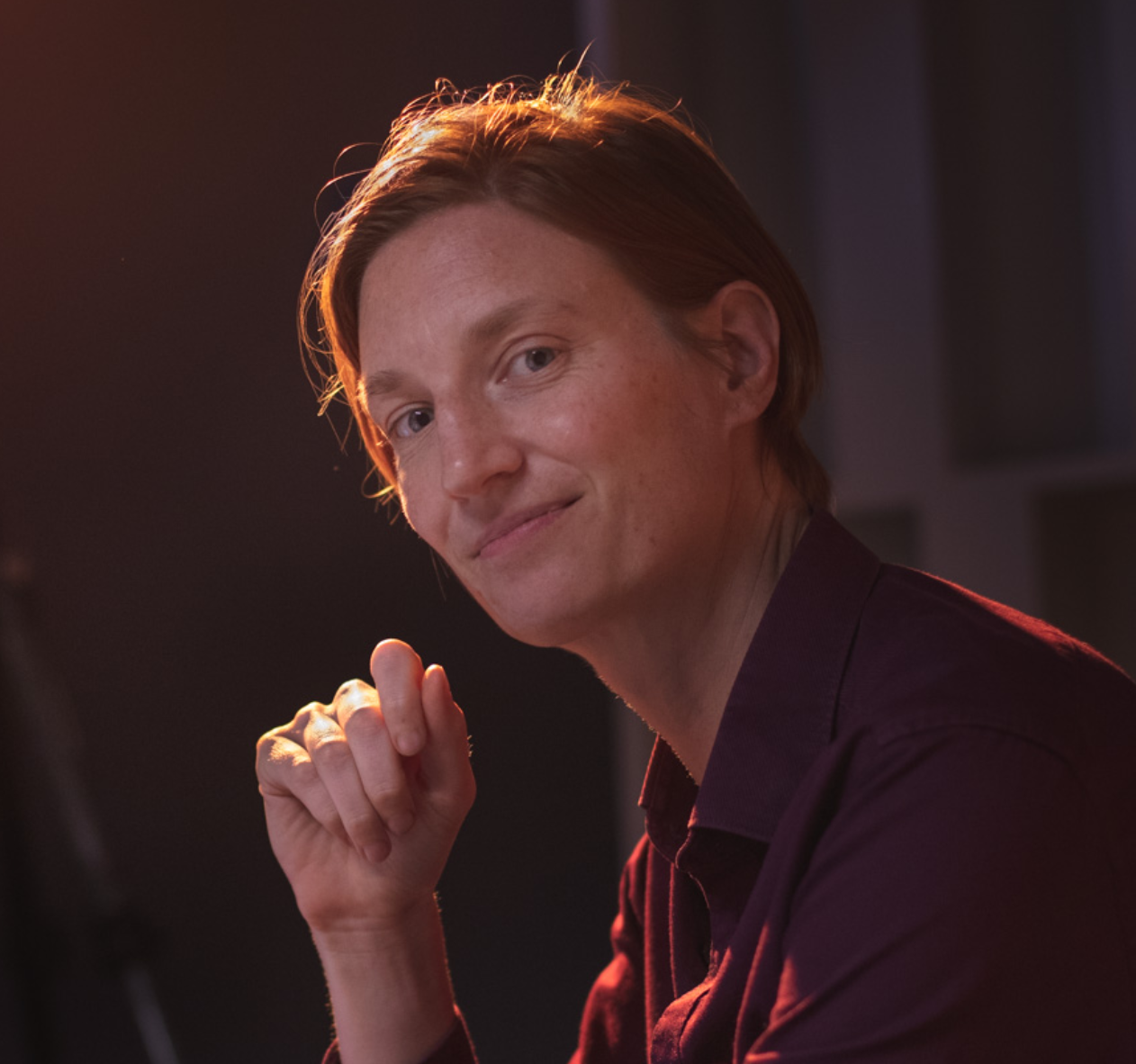}}]{Dorien Herremans} (M'12, SM'17) is an Assistant Professor at Singapore University of Technology and Design. In 2015, she was awarded the individual Marie-Curie Fellowship for Experienced Researchers, and worked at the Centre for Digital Music, Queen Mary University of London. Prof. Herremans received her PhD in Applied Economics from the University of Antwerp. After graduating as a business engineer in management information systems at the University of Antwerp in 2005, she worked as a Drupal consultant and was an IT lecturer at Les Roches University in Bluche, Switzerland. Prof. Herremans' research focuses on the intersection of machine learning/optimization and digital music/audio. She is a Senior Member of the IEEE and co-organizer of the First International
Workshop on Deep Learning and Music as part of IJCNN, as well as guest editor for Springer's Neural Computing and Applications and the Proceedings of Machine Learning Research. She was featured on the Singapore 100 Women in Technology list in 2021. 
\end{IEEEbiography}

\begin{IEEEbiography}[{\includegraphics[width=1in,height=1.25in,clip,keepaspectratio]{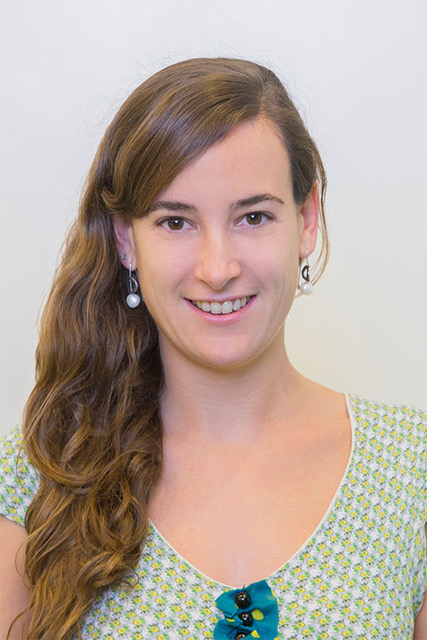}}]{Gemma Roig} is a professor at the computer science department in the Goethe University Frankfurt since January 2020. Before, she was an assistant professor at the Singapore University of Technology and Design. She conducted her postdoctoral work at MIT in the Center for Brains Minds and Machines. She pursued her doctoral degree in Computer Vision at ETH Zurich. Her research aim is to build computational models of human vision and cognition to understand its underlying principles, and to use those models to build applications of artificial intelligence.    
\end{IEEEbiography}

\vfill

\begin{IEEEbiography}[{\includegraphics[width=1in,height=1.25in,clip,keepaspectratio]{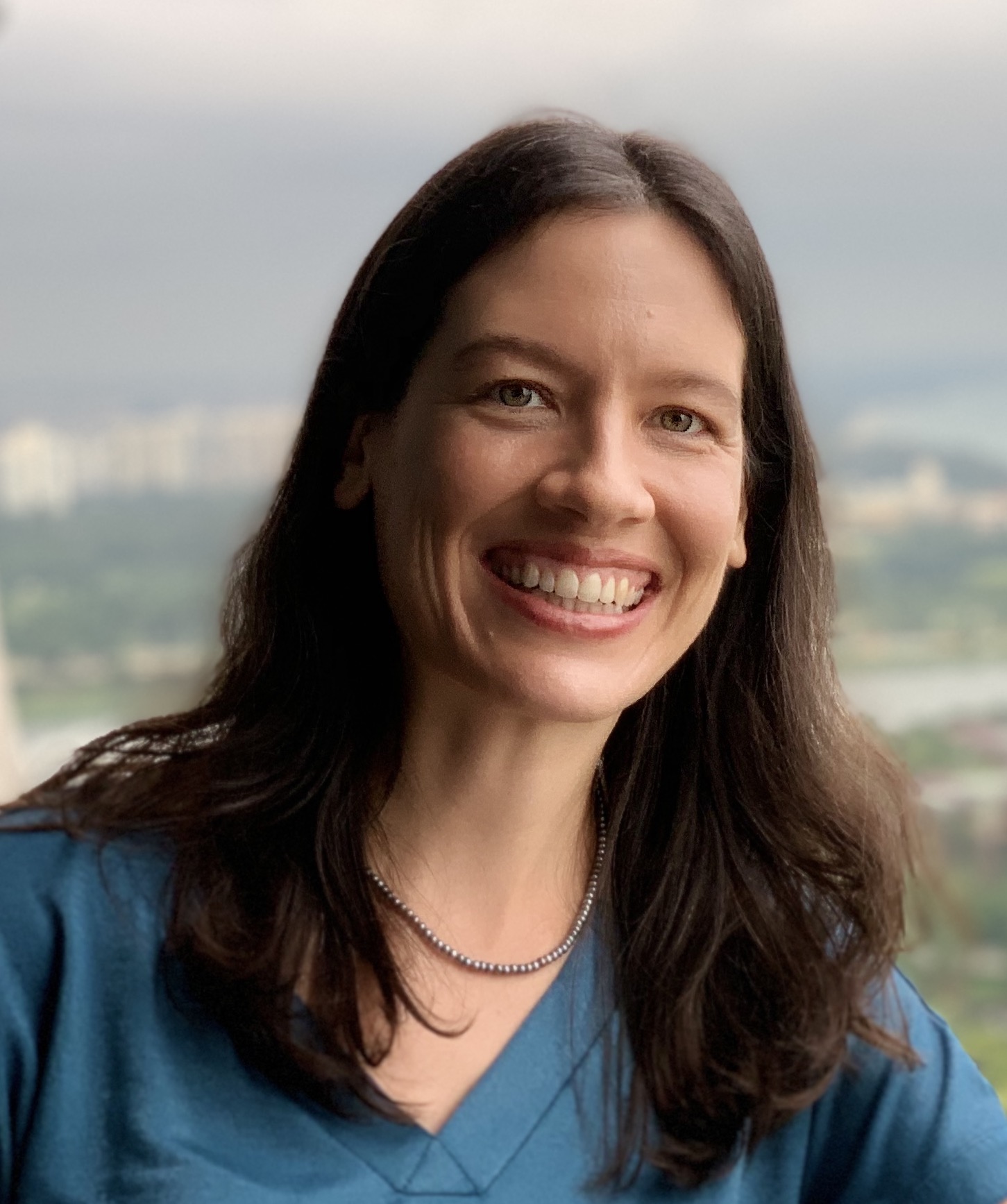}}]{Kat R. Agres} (M'17) is an Assistant Professor at the Yong Siew Toh Conservatory of Music (YSTCM) at the National University of Singapore (NUS), with a joint appointment at Yale-NUS College. Previously, she founded the Music Cognition group at the Institute of High Performance Computing, A$Q^{*}$STAR, in Singapore. Kat received her PhD in Psychology (with a graduate minor in Cognitive Science) from Cornell University in 2013, and holds a bachelor's degree in Cognitive Psychology and Cello Performance from Carnegie Mellon University. Her postdoctoral research was conducted at Queen Mary University of London, in the areas of Music Cognition and Computational Creativity. She has received funding from the National Institute of Health (NIH), the European Commission's Future and Emerging Technologies (FET) program, and the Agency for Science, Technology and Research (Singapore), amongst others, to support her research. Currently, Kat's research explores a range of topics including music technology for healthcare and well-being, music perception and cognition, computational modeling of learning and memory, and automatic music generation. She has presented her work in over fifteen countries across four continents, and remains an active cellist in Singapore.

\end{IEEEbiography}





\end{document}